# A robust particle detection algorithm based on symmetry

*Alvaro Rodriguez [†], Hanqing Zhang [†], Krister Wiklund [†], Tomas Brodin[‡], Jonatan Klaminder[‡], Patrik Andersson[§], Magnus Andersson[†,*]*

[†]Department of Physics, [‡]Department of Ecology and Environmental Science, [§]Department of Chemistry, Umeå University, 901 87 Umeå, Sweden

*Corresponding author: Magnus Andersson, Phone +46 90 786 6336, email: <u>magnus.andersson@umu.se</u>,

## Abstract

Particle tracking is common in many biophysical, ecological, and micro-fluidic applications. Reliable tracking information is heavily dependent on of the system under study and algorithms that correctly determines particle position between images. However, in a real environmental context with the presence of noise including particular or dissolved matter in water, and low and fluctuating light conditions, many algorithms fail to obtain reliable information. We propose a new algorithm, the Circular Symmetry algorithm (*C-Sym*), for detecting the position of a circular particle with high accuracy and precision in noisy conditions. The algorithm takes advantage of the spatial symmetry of the particle allowing for subpixel accuracy. We compare the proposed algorithm with four different methods using both synthetic and experimental datasets. The results show that *C-Sym* is the most accurate and precise algorithm when tracking micro-particles in all tested conditions and it has the potential for use in applications including tracking biota in their environment.

# 1. Introduction

Tracking micro-particles with computer-enhanced video microscopy is a common technique in biophysics and micro-fluidics.[1,2] By monitoring the movement of a particle using: fluorescence microscopy,[3] brightfield microscopy[4] or darkfield microscopy;[5] important information of the system under study can be revealed. Biological systems that have been investigated using these methods are for example; mechanical properties of polymers,[3,4] diffusion of individual proteins,[6] dynamic properties of DNA and interactions of DNA with different molecules.[7,8] Tracking micro-particles is also common when designing micro-fluidic devices.[9]

To obtain reliable data of motion, it is important that particle positions are accurately determined.[10] Accuracy is limited by the detection algorithm[11,12] and the microscope spatial resolution, which is determined by the quality of the optics and the wavelength of the light. To improve accuracy several algorithms have been developed during the years: e.g., Center-of-Mass (*CoM*),[13] Gaussian fitting, [14] Cross-Correlation (*XCorr*),[15] quadrant interpolation (*QI*)[16] and the Hough transform (*CHT*).[17] Though these algorithms are commonly used, they all have important limitations.

The *CoM* algorithm locates particle centers by selecting an intensity threshold value and generating a binary mask to evaluate the pixel positions of interest. This algorithm is fast and simple, but very sensitive to shot noise and light fluctuations.[18] The Gaussian fitting method is based on approximating the point spread function (the spatial intensity distribution) of a particle with Gaussian functions. The main limitations of this algorithm are that it is sensitive to defocus and that particles much larger than the wavelength of light cannot accurately be described by a Gaussian distribution.[12] The *XCorr* algorithm is based on locating objects by correlating different radial profiles to find the particle center of symmetry.[19] This algorithm is sensitive to noise and interference patterns from surrounding particles. The *QI* algorithm takes advantage of the circular geometry of non-diffraction-limited objects and uses image interpolation to achieve subpixel accuracy. *QI* requires, however, an accurate initial estimation of the particle position to perform well. Finally, the *CHT* algorithm is commonly used for detecting circular patterns and several modifications have been made of the algorithm to make it faster and more robust against noise; e.g., edge-drawing circles[20] or isosceles triangle circle detection.[21] The main advantage of *CHT* is that it is capable of handling occlusions. However, it is still more sensitive to noise than algorithms not based on edge features, and its accuracy is dependent on the size of the circular pattern.

To improve the accuracy and precision of particle detection we present a new approach, the Circular Symmetry (*C-Sym*) algorithm. The algorithm uses correlation analysis to determine the degree of symmetry. We hypothesize that, by using spatial symmetry, the particle position can be more accurately determined even in the presence of significant noise. *C-Sym* subsequently employs spatial filtering, piecewise Hermite interpolation and polynomial fitting to achieve subpixel accuracy and improve robustness. We evaluated the performance of *C-Sym* using: synthetic images that simulate spherical particles in different conditions;

and experimental data of micro-spherical particles in a bright field microscope. The results of *C-Sym* algorithm are compared with that of *CHT*, *CoM*, *XCorr* and *QI* algorithms. The results show that *C-Sym* has better accuracy and precision, especially when tracking particles in environment with noise.

## 2. Materials and Methods

### 2.1. Design of the experiment with synthetic images

To evaluate the accuracy and precision, defined as the magnitude of the error and the spread of the error, of particle detection we used computer generated (synthetic) images. This allowed us to have control over the experimental parameters without the errors introduced by experimental equipment; e.g., imaging system.[22,23]

A preliminary analysis indicated that particle size and noise level are variables with a strong impact on particle position estimation. Thus, in our experiments we independently tested both variables by generating particles with a known radius and then exposed the images to white-noise using a zero mean Gaussian distribution. We generated 1000 test simulations for each particle size and noise level, using the signal-to-noise-ratio (SNR) defined as,

$$SNR = \frac{I_{max} - I_{min}}{4\sigma} - 1, \tag{1}$$

where $I_{max}$ and $I_{min}$ are the maximum and minimum intensities of the image, and $\sigma$ is the standard deviation (SD) of the noise signal in the image. Examples for different SNR are shown in Figure S1.

Additionally, we identified several other variables that slightly influenced the particle position estimation and could act as a source of bias in the analysis, i.e., particle structure, ground truth position, estimated initial position, and background color. The particle structure, i.e., the visual appearance in the image, can vary considerably according to the experimental setup and the nature of the particle itself. This can significantly change the accuracy of the algorithm used for position estimation. We generated different patterns representing micro-particles, which varied from a simple "spot" to diffraction patterns, see Figure 1. The diffraction patterns were generated using a diffraction profile extracted from a real particle and by introducing random variations such as; inversion of color, profile scaling and stretching. To avoid bias when generating synthetic images, we set the ground truth position as a random floating-point pixel position. For the initial estimation of particle position, we introduced a random error up to 2 pixels to simulate the labeling or segmentation error. Finally, the background intensity of each image was randomly selected from values between 0.25 and 0.75 representing typical test conditions. Randomized synthetic images (512x512 pixels) for each particle size with the above mentioned variables were generated. In addition, we added Gaussian noise with zero means and variance $s^2$ to the images. A schematic diagram showing the generation of test images of certain particle size and noise level is shown in Figure 1. Additional reference particles are shown in the Supporting Material in Figure S2.

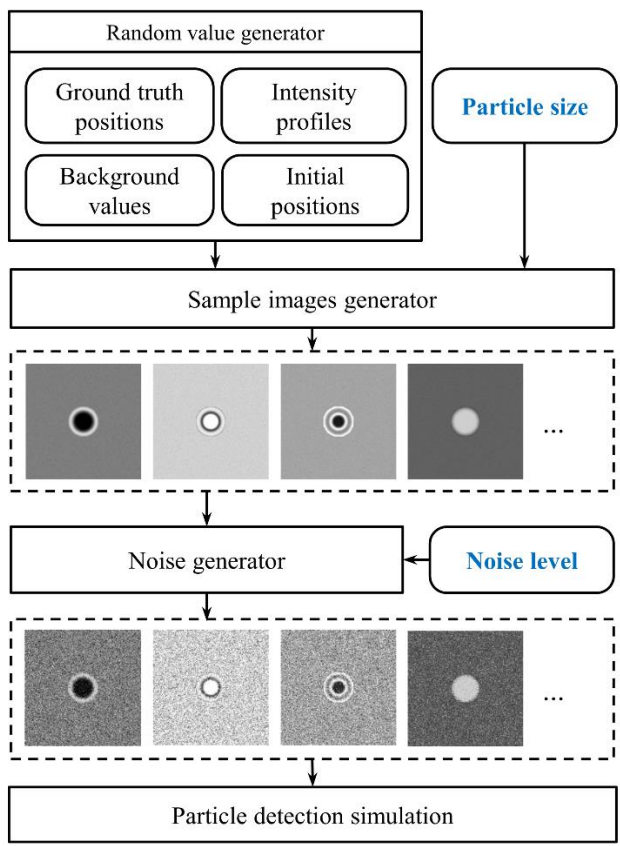

**Figure 1** The block diagram of generating sample images. Two major variables, particle size and noise level, and other relevant variables were used to create synthetic images.

In summary, to avoid biased results we conducted 1000 trials for each particle size and noise level using: randomized background colors, particle patterns and particle positions with a constant region of interest (ROI) size. Each algorithm was evaluated in terms of its accuracy and precision using the mean and standard deviation of the Euclidean distance between the estimated position of the particle and the ground truth.

## 2.2.    Design of the experiment with micron-sized particles

To validate *C-Sym* we also generated experimental data by oscillating a micron-sized particle in a bright field microscope and evaluated the error of amplitude (peak-to-peak displacement distance) by comparing the tracked position with the real piezo-stage motion. We prepared our sample using silica micro-spheres (Catalog Code SS04N, Manufacturer Lot Number: 7829 – Bang Laboratories) with a diameter of 2.0 µm suspended in a solution of Milli-Q water. Micro-spheres were immobilized to a cover slide by drying 10 µl of the suspension in room temperature. The samples were then placed in a microscope (Olympus, IX71), modified for optical tweezers and flow chamber experiments, and imaged using a water immersion objective (UPlanSApo 60x, Olympus).[24] A representative image of a micro-sphere is shown in Figure 2a. The micro-spheres were moved by a piezo stage with sub-nm resolution (PI-P5613CD, Physik Instruments). We used a CCD camera (C11440-10C, HAMAMATSU, 8 bit) with a conversion factor (1 px = 84.7 ± 0.5 nm, mean ± standard error (SE) on the mean, and the standard deviation (SD) was 1.6 nm), to record the motion.

To obtain interpretable results, we compared the displacement of the piezo stage with the displacement obtained by fitting detected particle positions to a sinusoidal function

$$f(i) = a\sin(2\pi i / b + 2\pi / c) + d, \qquad (2)$$

where $i$ is the frame number, $a$ is the amplitude of displacement in pixels, $b$ is the period, $c$ represents the phase and $d$ is an offset. In this case, $b$ is a constant defined by multiplying the cameras sampling speed and the piezo stage oscillation period.

(a)

(b)

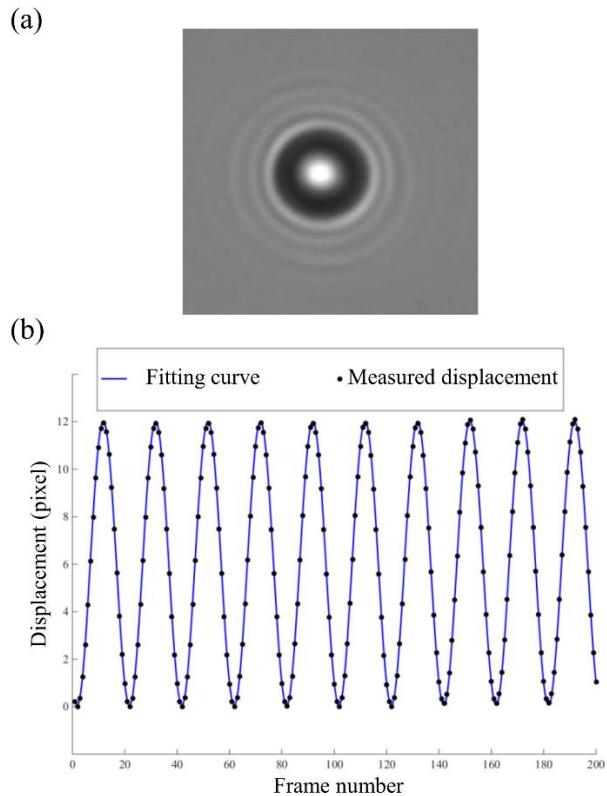

**Figure 2** (a) Example of an image from the experiment conducted with a 2.0 µm silica micro-sphere immobilized to a cover slip. (b) Measured displacement of the particle with the proposed *C-Sym* algorithm (marked as circles) and the ground truth sinusoidal function for the first 200 frames of the experiment (solid line).

To transform the displacement in pixels to real space distance, an additional calibration process is required. In an ideal projection system, like a pin-hole camera, distances from the image plane can be transformed to the real plane of interest using a multiplication factor. However, when using an imaging system a more complex calibration model is required to deal with plane misalignment[25] or lens aberrations,[26] and other systematic errors of the instruments.[27] For simplicity, we modelled the image to real-space-projection as a parametric polynomial function. Based on our results, we choose a 5th degree parametric polynomial function depending on the amplitude of the displacement expressed as,

$$D = p_6 a^5 + p_5 a^4 + p_4 a^3 + p_3 a^2 + p_2 a + p_1, \tag{3}$$

where $D$ is the real amplitude of the particle displacement in nm, and $p_i$ are the calibration parameters. We used 11 video sequences with a known $D$. Thereafter, all algorithms were used to measure the amplitude $a$ from the videos, and the calibration parameters $p_i$ were calculated by minimizing the projection error (the distance between the projection of measured amplitude $a$ and $D$). By applying this approach, the resulting projection error was <1% and the correlation with the ground truth was statistically significant (t-test, p<0.05) and with a correlation value >0.999.

## 2.3.    Design of the experiment with tethered particles

Using bright field experimental data of supercoiled tether DNA, we tested the accuracy of *C-Sym* in comparison to the other algorithms. The video data was taken from the studies in reference.[28] In these experiments, supercoiled DNA (pSB4312) was immobilized at one end to a coverslip using PNAs while the other end was attached to a 0.51 µm Streptavidin-coated micro-sphere (cat. no. CP01N, Bangs Laboratory). The motion of a micro-sphere was recorded at 225 Hz in an inverted microscope (Model No. DM IRB, Leica) with a high-speed camera (Pike F100B, Allied Vision Technology, conversion factor of 49.97 nm per pixel). The length of the time series were optimized to 30 seconds according to an Allan variance analysis of the setup.[29] Representative images (62x62 pixel ROI) are shown in Figure 3a.

In these experiments, no ground truth of the particle position was available since a micro-sphere will move stochastically due to the Brownian motion. However, since the appearance of the microsphere in each frame, Figure 3a, only marginally changed from frame to frame because the motion is more or less constrained in a plane, the algorithms were applied to each frame to extract particle positions. A ROI of fixed size was applied around the detected particle in each frame, see Figure 3b. The ROI from frame $i$ was thereafter correlated with ROI$_{i+1}$ resulting in a number between 0 and 1, where the latter corresponds to perfect correlation. The consecutive correlation time series of an experiment is shown in Figure 3c. Thus, if an algorithm is able to accurately predict the position of the particle, the correlation value should be close to 1. Note that this does not provide a direct measurement of accuracy for the algorithms; however, it measures the similarity of ROIs with the particle positioned in the center and therefore reveals the accuracy and precision of particle detection.

In the original video sequences, we estimated the mean SNR of frames to be 10. To test the algorithms for robustness against noise, we applied to each frame white-noise with zero mean and different variance, σ, to obtain the mean SNRs in the range from 0.1 to 10. At each noise level we evaluated each algorithm using the relative difference of correlation values. The particle position in a noisy image were found by the algorithms and used in the original image as $(x_i, y_i)$, see procedure in Figure 3.

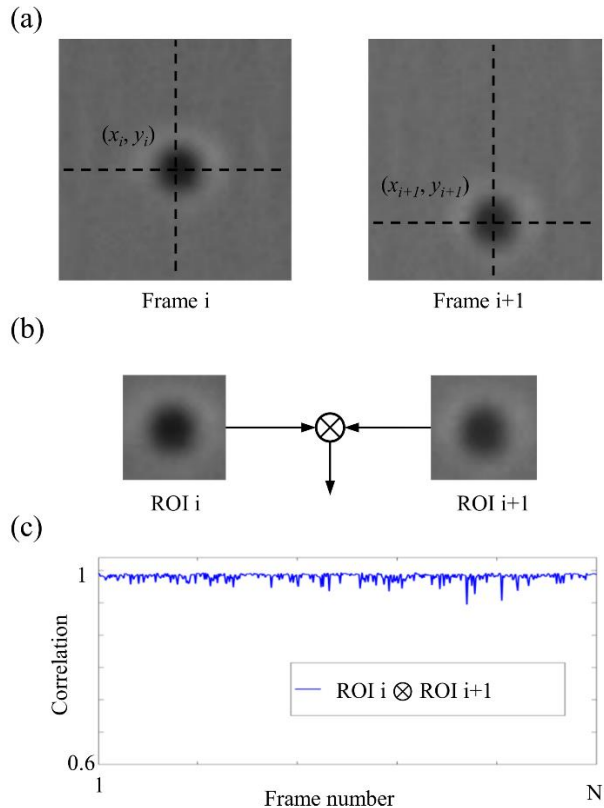

**Figure 3** Evaluation of the performance of the algorithms using a tethered micro-sphere. (a) The algorithms are used to locate the micro-sphere position in each frame. (b) The micro-sphere is extracted from every frame using a constant sized ROI centered on the detected position. Consecutive ROIs are correlated as denoted by the operator ⊗. (c) The correlation for each frame number. An algorithm with poor precision will give a low correlation value, thus, the evolution of correlation is an indicator of the stability and robustness of the algorithm used to locate the micro-sphere.

## 3.  Central-Symmetry algorithm (*C-Sym*)

Accurately extracting the particle position from images can be problematic if the images have: a substantial amount of noise, changing background colors, or changing particle appearance. Our approach to handle this, is to use geometrical symmetry to find the particle position. A rough estimate of the particle position is first needed. This can be obtained using standard methods, e.g., manual selection, template matching, or segmentation, but the choice of method is not crucial for the performance and the final results. Since the algorithm consist of several steps, we present these steps in the workflow chart in Figure 4, and each step is described below.

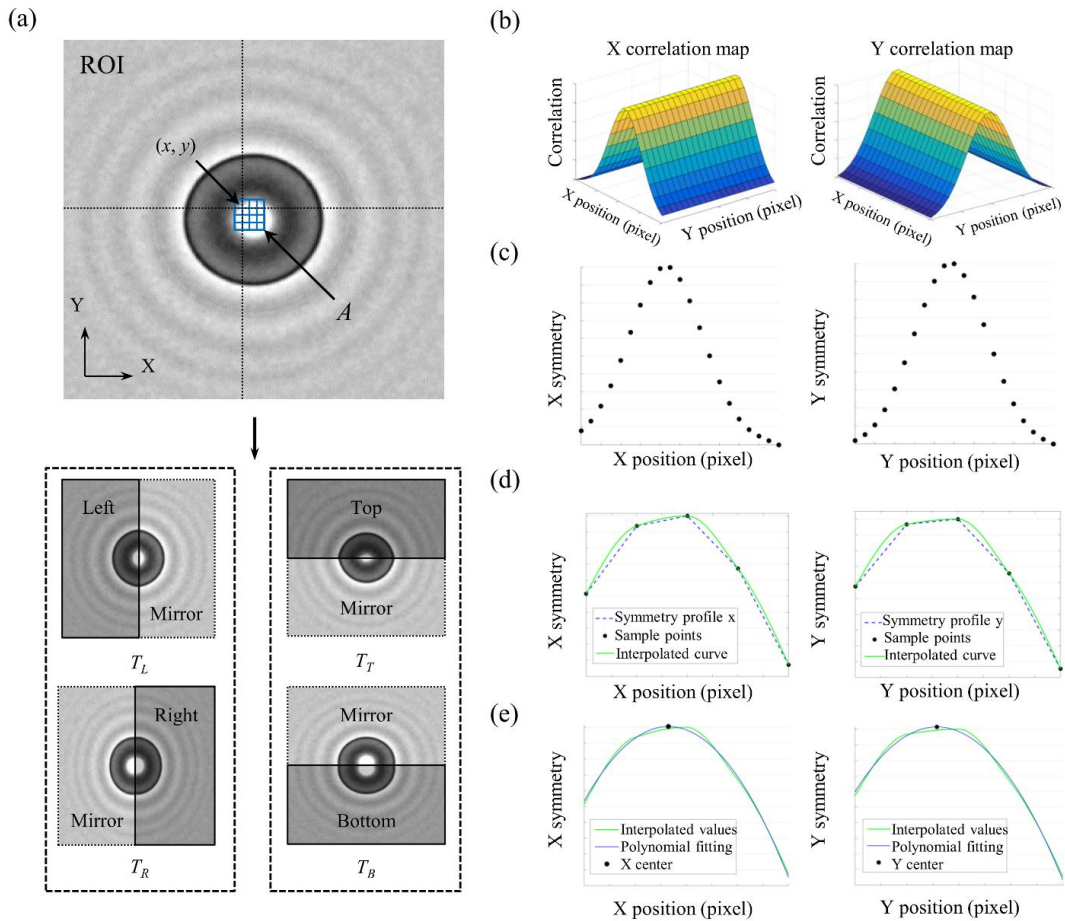

**Figure 4** The workflow of the C-Sym algorithm. (a) For candidate points (x,y) in a search area A, a region of interest (ROI) is defined and four templates of the particle are created, dividing the ROI horizontally and vertically and reconstructing the whole particle from each template. (b) Pairs of templates are used in Equation 4 and 5, to create the 3-D correlation maps, CorrX and CorrY. (c) 2-D symmetry profiles, SymX and SymY, are created from correlation maps using average filtering defined by Equation 6 and 7. (d) Symmetry profiles are interpolated using the Hermite algorithm. (e) Correlation centers are obtained from interpolated Symmetry profiles with polynomial fitting.

**(a) Template extraction:** a search area $A$, defined as a two dimensional array, is positioned around the initial center estimation, see Figure 4a. For each point $(x, y)$ in $A$, a ROI is created with a fixed size ($n \times n$) defined by the user. In general, this ROI should be slightly larger than the particle. The ROI is divided vertically at $x$, and a mirror image of each half is; created, flipped and concatenated to each half to form two templates denoted $T_L$ and $T_R$. A similar process is conducted horizontally at $y$ resulting in two new templates, $T_T$ and $T_B$, as shown in Figure 4a. The templates denoted $T$, are all functions of $x$ and $y$, (i.e., $T_L^{x,y}$), for clarity we choose not to write that dependency explicitly.

**(b) Symmetry measurement:** the 3D correlation map, $Corr_X(x, y)$, is derived from the template pair $T_L$ and $T_R$, representing the vertical symmetry of the particle at the candidate point $(x, y)$ in $A$. Likewise, the 3D correlation map $Corr_Y(x, y)$ is derived from $T_T$ and $T_B$, representing the horizontal symmetry of the particle at $(x, y)$. The correlation maps are defined as

$$Corr_X(x, y) = \frac{\sum_{i,j}^{n \times n}\left(\left(T_L(i,j) - \mu_L\right)\left(T_R(i,j) - \mu_R\right)\right)}{\sigma_L \sigma_R}, \tag{4}$$

$$Corr_Y(x, y) = \frac{\sum_{i,j}^{n \times n}\left(\left(T_T(i,j) - \mu_T\right)\left(T_B(i,j) - \mu_B\right)\right)}{\sigma_T \sigma_B}, \tag{5}$$

where $T_L$, $T_R$, $T_T$, $T_B$ represent the four templates of size $n \times n$ built around the coordinates $(x, y)$; and $\mu$ and $\sigma$ are the average and standard deviation of the corresponding template.

**(c) Dimensional filtering:** To increase the robustness against noise an average filter is applied in each 3D correlation map. This reduces the correlation maps into two 2D symmetry profiles, $Sym_X$ and $Sym_Y$, as shown in Figure 4c and defined by

$$Sym_X(x) = \frac{1}{N}\sum_{y=1}^{N} Corr_X(x, y), \tag{6}$$

$$Sym_Y(y) = \frac{1}{N}\sum_{x=1}^{N} Corr_Y(x, y), \tag{7}$$

$Sym_X$ and $Sym_Y$ will have a maximum when $T_L$ and $T_R$, and $T_T$ and $T_B$ are identical. This maximum represents the center of symmetry of the particle, thus providing the center position.

**(d) Piecewise Hermite interpolation:** To improve the accuracy, we use the Hermite piecewise algorithm,[30] which allows us to increase the resolution of the symmetry profiles and to find the maximum in $Sym_X$ and $Sym_Y$ with subpixel accuracy.[31] The advantage of using piecewise Hermite interpolation in our algorithm is shown and discussed in the Supporting Material.

In the interpolation process for $Sym_X$, the discrete points $x_1, x_2, ..., x_n$ are used to create a set of third-degree polynomials defined as,

$$p_k^x(t) = (2t^3 - 3t^2 + 1)x_k + (t^3 - 2t^2 + t)m_k + (-2t^3 + 3t^2)x_{k+1} + (t^3 - t^2)m_{k+1}, \qquad (8)$$

where $t$ are values in the interval $[x_k, x_{k+1}]$; $m_k$ is the slope of the tangent line of $Sym_X$ at $x_k$ and $m_{k+1}$ is the slope of the tangent line of $Sym_X$ at $x_{k+1}$ respectively. The same procedure is applied to the Y axis. The results of this process are two continuous curves with a continuous first derivative which passes through the sampled values of $Sym_X$ and $Sym_Y$ respectively. These continuous curves are denoted as $S_X$ and $S_Y$.

**(e) Peak analysis:** The continuous curves $S_X$ and $S_Y$ represent the symmetry of the particle in $A$ and their peaks represent the X and Y coordinates of the particle respectively. To find these peaks, $2^{nd}$ degree polynomials are fitted to $S_X$ and $S_Y$ using 500 samples located in the neighborhood of the discrete peak and with a sample interval of 0.01 pixels. Each polynomial equation was solved using a Vandermonde matrix. Once the polynomial is obtained, the center is calculated using the equation defined as,

$$pol(x) = q_1 x^2 + q_2 x + q_3, \qquad (9)$$

where $q_i$ are the polynomial parameters. The center of the particle in the X-axis, $c_x = -q_2 / 2q_1$, is the peak of the fitted polynomial. The same procedure is used to locate the center of the particle in the Y-axis. When the Hermite interpolation is omitted, the $2^{nd}$ degree polynomials are fitted to $Sym_X$ and $Sym_Y$ using 5 discrete samples in the neighborhood of the peak.

# 4. Results and Discussion

The performance of the *C-Sym* algorithm in the particle detection experiments with simulated and real particles are outlined in detail below. Note that the same datasets of particles were used to compare *C-Sym* with four state of the art algorithms: *CHT*, *CoM*, *XCorr* and *QI* and that details of these latter algorithms are described in the Supporting Material S2.

## 4.1. Experiment with synthetic images

In line with our hypothesis, *C-Sym* showed better accuracy and precision, especially for noisy images. The evaluation was performed by using 110 000 simulated particle images with a resolution of 512x512 pixels and by varying the SNR from 0.1 to 100 in 11 steps. The particle radii varied from 10 to 100 pixels in 10 steps in these images. Estimated errors for each fixed particle size and SNR are shown in Figure 5. To further illustrate the effect of SNR and particle radius, we chose three selections of the results with a constant particle radius at 10, 50 and 100 pixels as shown in Figure 6.

Both *CHT* and *CoM* showed lower accuracy than the other algorithms for SNR lower than 50, i.e. for images with more noise. However, *XCorr*, *QI*, and *C-Sym*, all performed well down to a SNR of 1. Here, *QI* was slightly better than *XCorr*, whereas *C-Sym* was better than both the others and able to provide high accuracy, 0.2 pixels mean error, at a SNR of only 0.1.

Particle size had a more complex effect on the result; implying for example that bigger particles are more corrupted by noise. The response of *CHT* showed high error values for some particle sizes even in low noise levels and *CoM, XCorr* and *QI* generated low accuracies for particles smaller than 20 pixels in radius. However, *C-Sym*, does not show this limitation, achieving a maximum mean error close to 0.1 pixel with particles of 10 pixel radius.

The precision of each algorithm was evaluated using the Standard Deviation (SD) of error in particle position (Figure 7) and Figure 8 shows three selections of the results with a constant particle radius at 10, 50 and 100 pixels. The precision results were very similar to the accuracy results. In particular, precision of *CHT* and *CoM* decreased significantly with noise, whereas *XCorr* and *QI* showed the best precision with large particles and low noise levels. For these conditions, *C-Sym* showed similar precision to *QI,* however, *C-Sym* outperformed *QI* for small particles and high noise levels.

In conclusion, we have shown that *CHT* is, in general, inaccurate when measuring particle positions and that *CoM* has the disadvantage of being very sensitive to noise. Yet, *XCorr* and *QI* perform better, where *QI* slightly outperformed *XCorr*. This finding was expected, because *QI* uses *XCorr* in its first step and thereafter refines the results. These results are also consistent with previous findings.[12] However, our data suggest that *QI* is not significantly better than *XCorr*. Overall, *C-Sym* showed the best results, achieving generally similar or better accuracy and precision than the compared algorithms, while still being able to measure smaller particles and being more robust against noise.

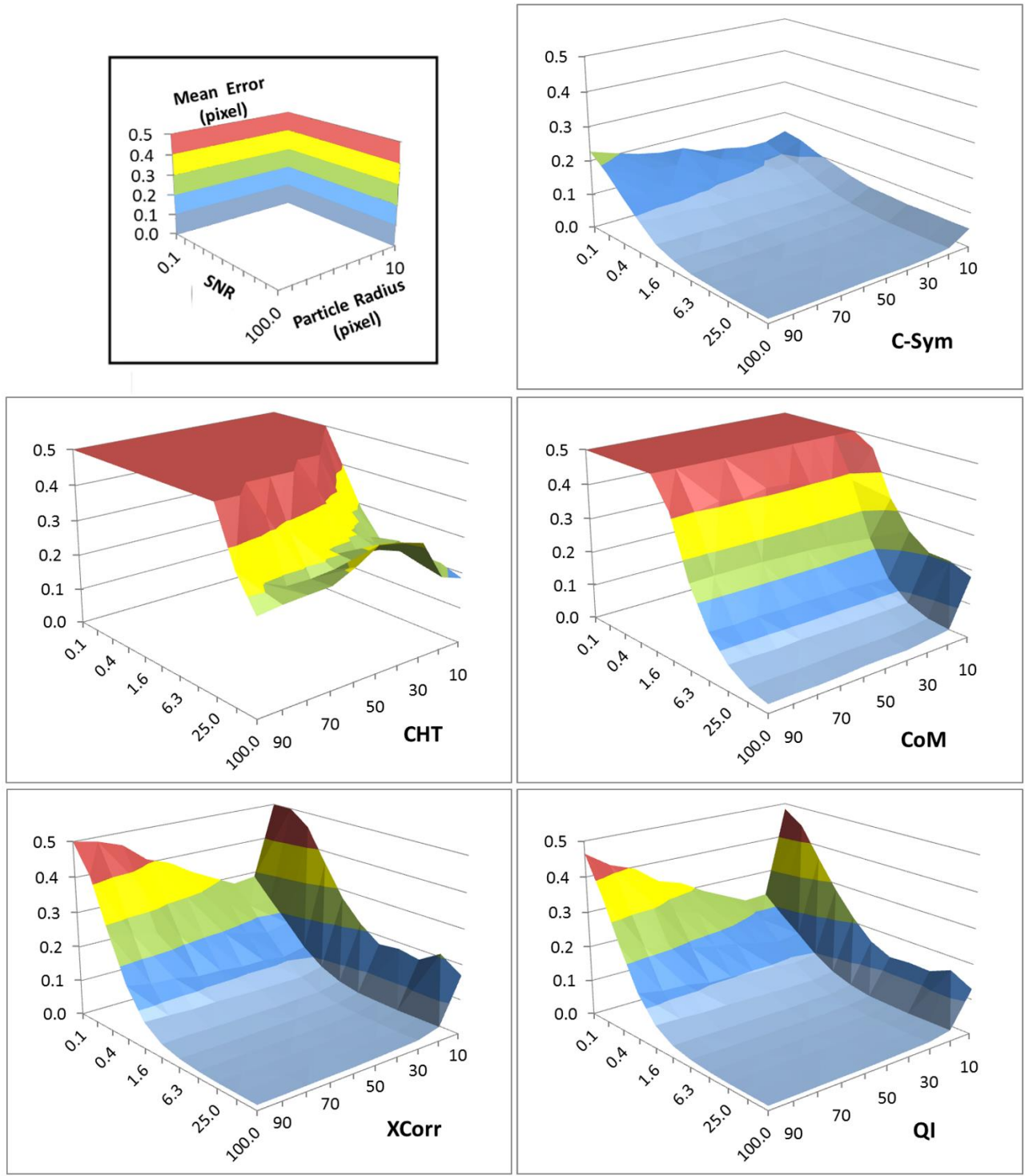

**Figure 5** Mean error in the center position of the particles measured using: C-Sym, CoM, CHT, XCorr and QI algorithms for different particle radius and noise levels. The upper left panel shows the scales and label of each axis. A low value of the SNR indicates noisy images.

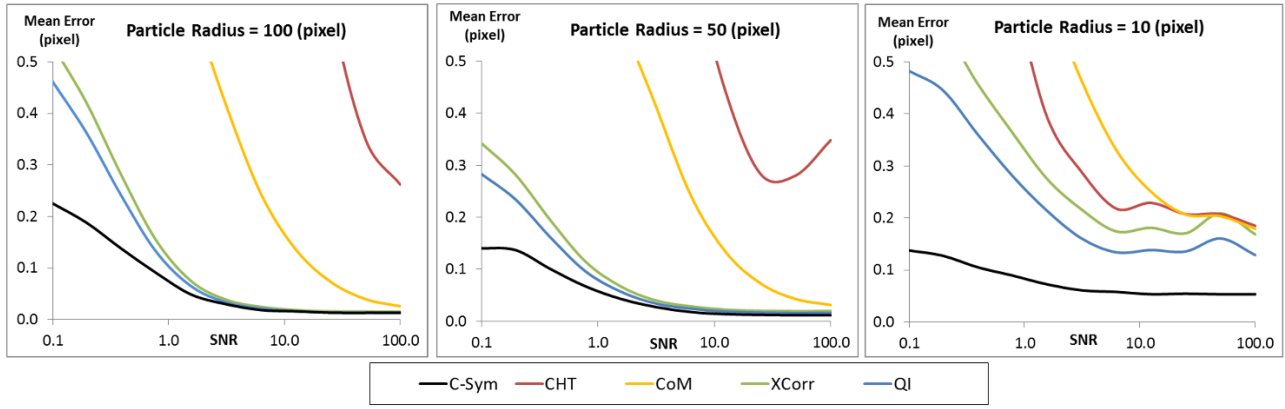

**Figure 6** Comparison of the mean error in the center position of a particle of constant radius for different SNR values.

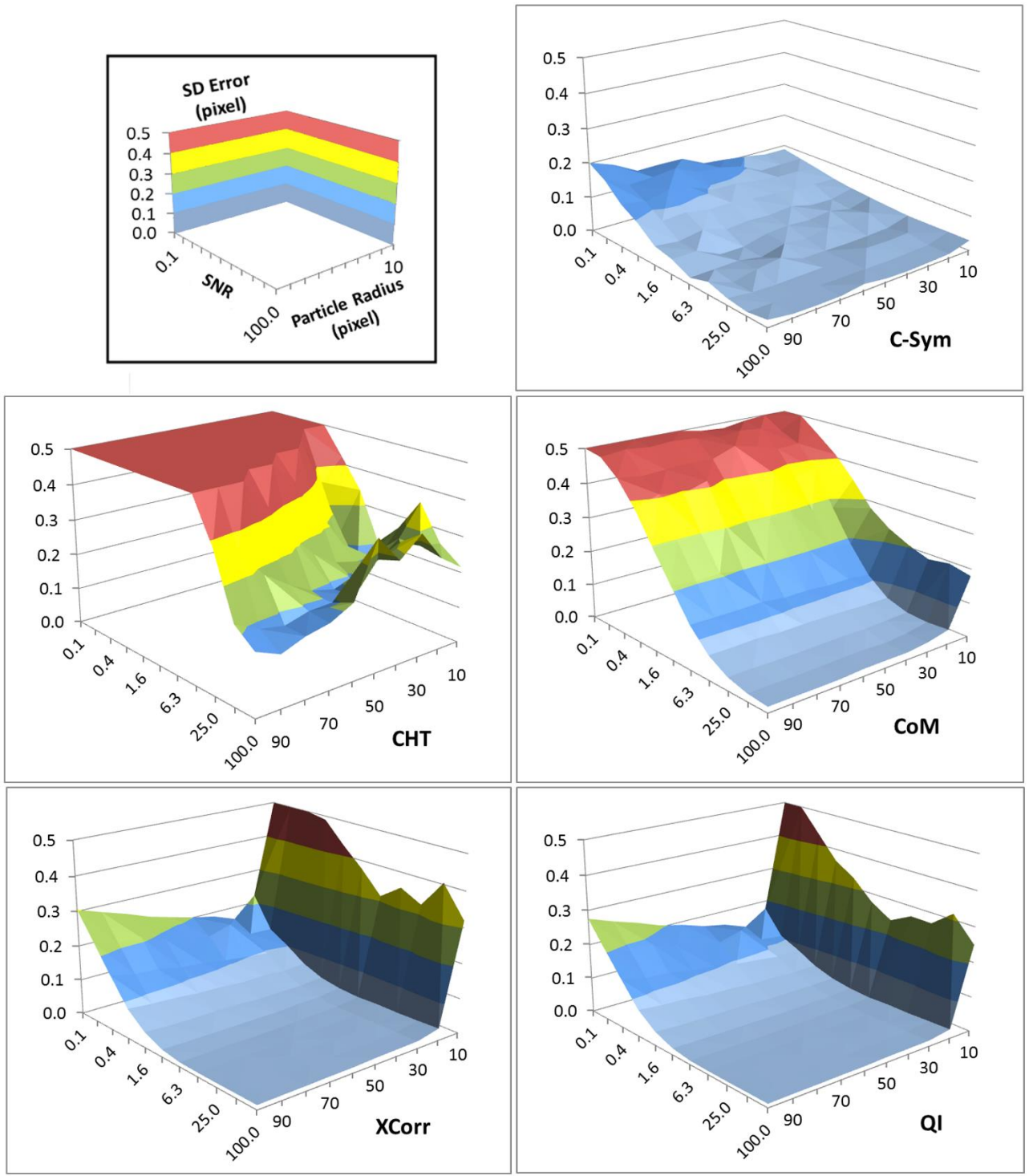

**Figure 7** Standard Deviation of the error in the position of the center of particles measured with C-Sym, CHT, CoM, XCorr and QI, algorithms according to the particle radius and the noise level.

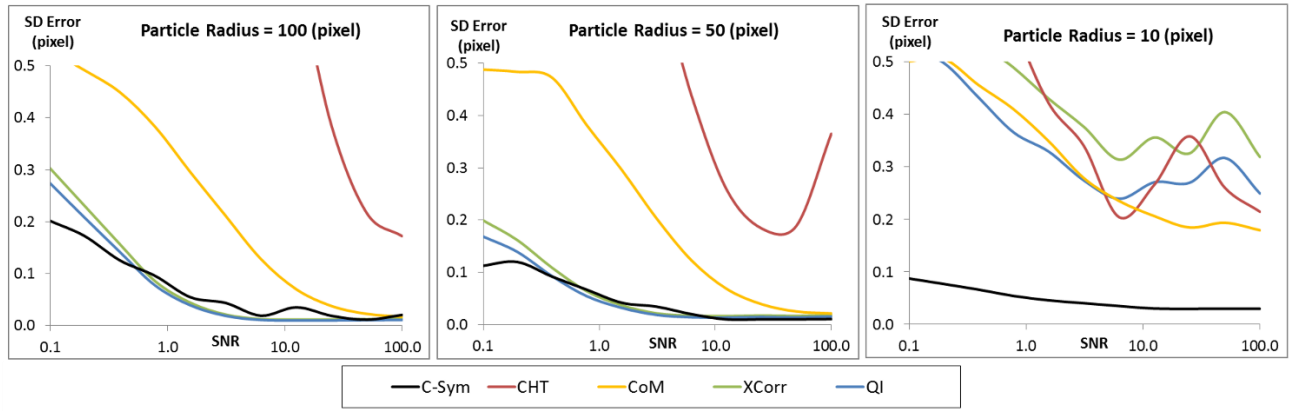

**Figure 8** Comparison of how the standard deviation of the error in the position of the center of particles measured with different algorithms change with noise level while using a constant particle radius of 100 (left) 50 (center) and 10 (right) pixels.

### 4.2.    Experiment with micron-sized particles

To validate and compare the performance of *C-Sym* using real data, 40 video sequences with 16 000 frames acquired at 500 Hz of 2 µm micro-spheres attached to a glass surface were analyzed. A micro-sphere was oscillated at 1 Hz with a sinusoidal function and varying peak-to-peak amplitude, from 1 to 900 nm. The estimated SNR of the video sequences is 50. The comparison of the aforementioned algorithms was done by using a relative amplitude error analysis (the absolute error divided by the amplitude of the displacement), as shown in Figure 9 and Table I.

The obtained results were generally consistent with previous finding.[13,15,16] *CHT* obtained the worst results, with an mean error >2 nm, and with a relative error >20% for smallest displacements. *XCorr* and *QI* performed better, but they still obtained an error >10% for the smallest displacements. The proposed *C-Sym* algorithm showed significantly better results, achieving an mean and standard deviation of error <1 nm, with a maximum relative error of 5%. Notably, these results do not support the statements in reference [12] which claims that *CoM* was unable to track particles in a similar scenario and that *XCorr* obtains large errors compared to *QI*.

The results are consistent with the synthetic experiments, and indicate that *C-Sym* is more stable and obtains smaller errors than the other algorithms. In this experiment *C-Sym* was the only algorithm able to measure amplitudes of nm order.

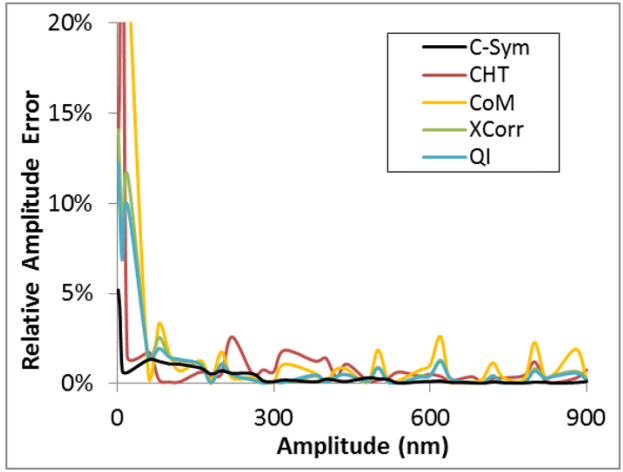

**Figure 9** Relative error of the amplitude of particle displacement with C-Sym, CHT, CoM, XCorr and QI, algorithms.

Table I: Absolute amplitude errors for the different algorithms in nm.

| Algorithm | Absolute Amplitude Error (nm) | | |
|-----------|------|------|---------|
|           | *Mean* | *SD* | *Maximum* |
| *C-Sym* | 0.71 | 0.45 | 1.57 |
| *CHT* | 2.43 | 2.19 | 9.66 |
| *CoM* | 3.55 | 4.57 | 18.30 |
| *XCorr* | 1.82 | 1.92 | 8.11 |
| *QI* | 1.67 | 1.69 | 7.59 |

### 4.3.    Experiment with tethered particles

Experiments using brightfield video sequences of a micro-particle attached to a coverslip through DNA strands were also conducted. In general, *CHT* were not able to locate the particle position, therefore this algorithm was excluded from the comparison. This was expected since the *CHT* requires a well-defined circular pattern; and this is not found in the images since the micro-particles show a diffuse Gaussian intensity distribution. All other algorithms performed well with a mean correlation >0.95 and a SD correlation <0.02 in all the SNR range as shown in Figure 10. As in our previous results, *C-Sym* performed overall better than the other algorithms, especially under high noise levels. Noticeably, *QI* showed slightly worse results than *XCorr*, and *CoM* performed better under high noise levels (SNR < 1) than both *XCorr* and *QI*.

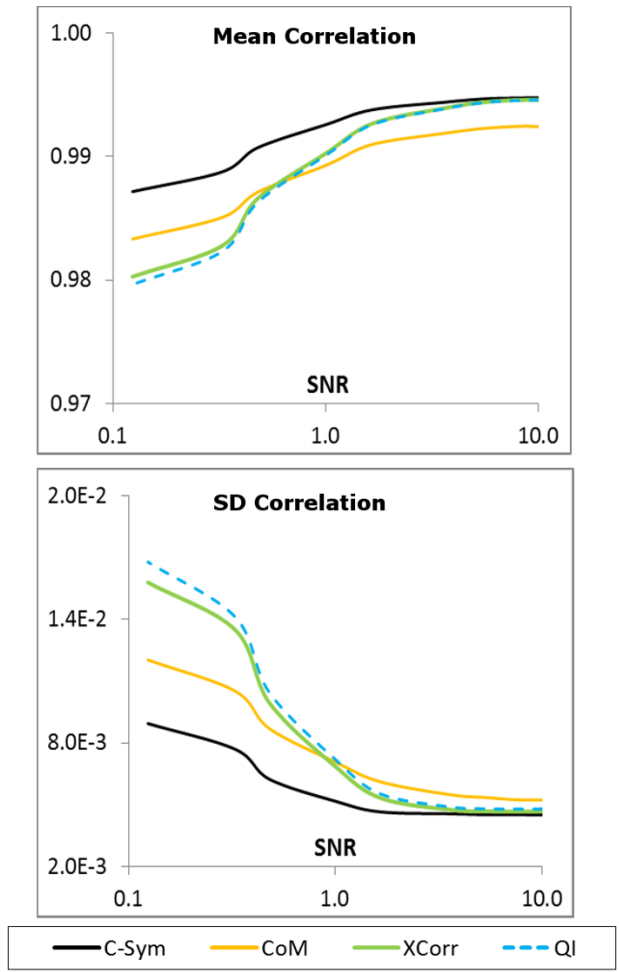

**Figure 10** Mean and SD correlation results in the experiment with tethered particles using C-Sym, CoM, XCorr and QI algorithms.

# 5. Conclusion

We propose a new algorithm, denoted the Circular Symmetry algorithm (*C-Sym*), for accurately locating the center of a circular particle. *C-Sym* uses the symmetry of the particle to achieve robust sub-pixel accuracy, capable of handling general circular patterns obtained from images. The strength of the algorithm is that even in noisy conditions, useful information of the particle is kept in the spatial distribution of the symmetry feature.

We compared the algorithm with other state-of-the-art methods: Circular Hough Transform *(CHT)*, Center-of-Mass *(CoM)*, Cross-Correlation *(XCorr)* and Quadrant Interpolation *(QI)* algorithms using synthetic and experimental images. The results show that *C-Sym* is more robust and achieves a higher accuracy and precision when measuring particle positions for a wide range of noise levels.

The robustness against noise in images is in particularly useful when studying systems with low light conditions. For example, fast processes that require short shutter times, and optical systems with high f-numbers. In addition, studying fast moving particles in micro-fluidic environments are often subjected to low SNR and particle sizes may vary due to displacements in depth. Therefore, *C-Sym* is expected to be a new useful algorithm for various biophysical systems, giving a reliable estimation of particle location, even though the environment is noisy or the resolution of particles is low.

In this work we have shown that by using 2D spatial symmetry features of micro-particles we can accurately determine their center position. A future work is to expand this symmetry algorithm and investigate if it can find the bilateral axis of symmetry, i.e., the center line of a geometrical object with identical left and right sides. This can prove very useful in several biological applications, e.g., when classifying animals, handling occlusions, and when determining the direction of movement of fish or insects in biological assays.

## Acknowledgements

This work was supported by the Swedish Research Council (2013-5379) and from the Kempe foundation to M.A.

# A robust particle detection algorithm based on symmetry

*Alvaro Rodriguez[†], Hanqing Zhang[†], Krister Wiklund[†], Tomas Brodin[‡], Jonatan Klaminder[‡], Patrik Andersson[§], Magnus Andersson[†,*]*

[†]Department of Physics, [‡]Department of Ecology and Environmental Science, [§]Department of Chemistry, Umeå University, 901 87 Umeå, Sweden

*Corresponding author: Magnus Andersson, Phone +46 90 786 6336, email: magnus.andersson@umu.se,
**Keywords:** particle tracking, microscopy, fluidics, image processing, micro-sphere

# Supporting Material

## S1: Design of the experiment with synthetic images

To evaluate the accuracy of particle detection we used synthetic images that varied from simple "spots" to complex diffraction patterns. These particles were generated using a diffraction profile extracted from a real particle and by introducing random variations such as; inversion of color, profile scaling and stretching. Also, the background intensity of each image was randomly selected. In this way, we generated for the experiments particles with random appearances and backgrounds and with a selected particle size. Examples of a particle with different SNR levels are presented in Figure S1. Examples of generated synthetic images with different particle sizes and patterns are presented in the Figure S2.

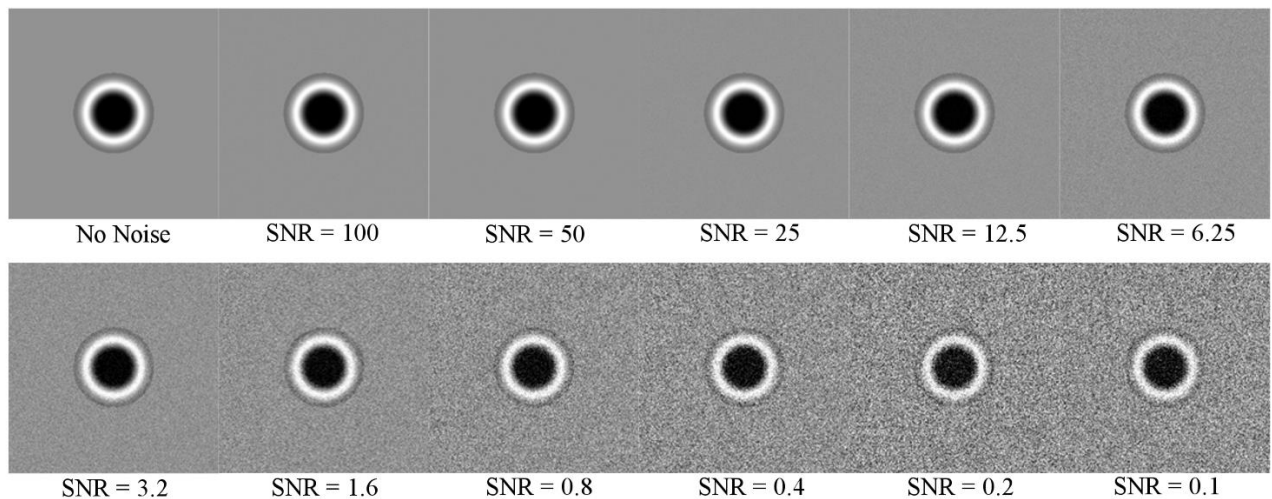

Figure S1. Images of a synthetic particle at different SNR levels.

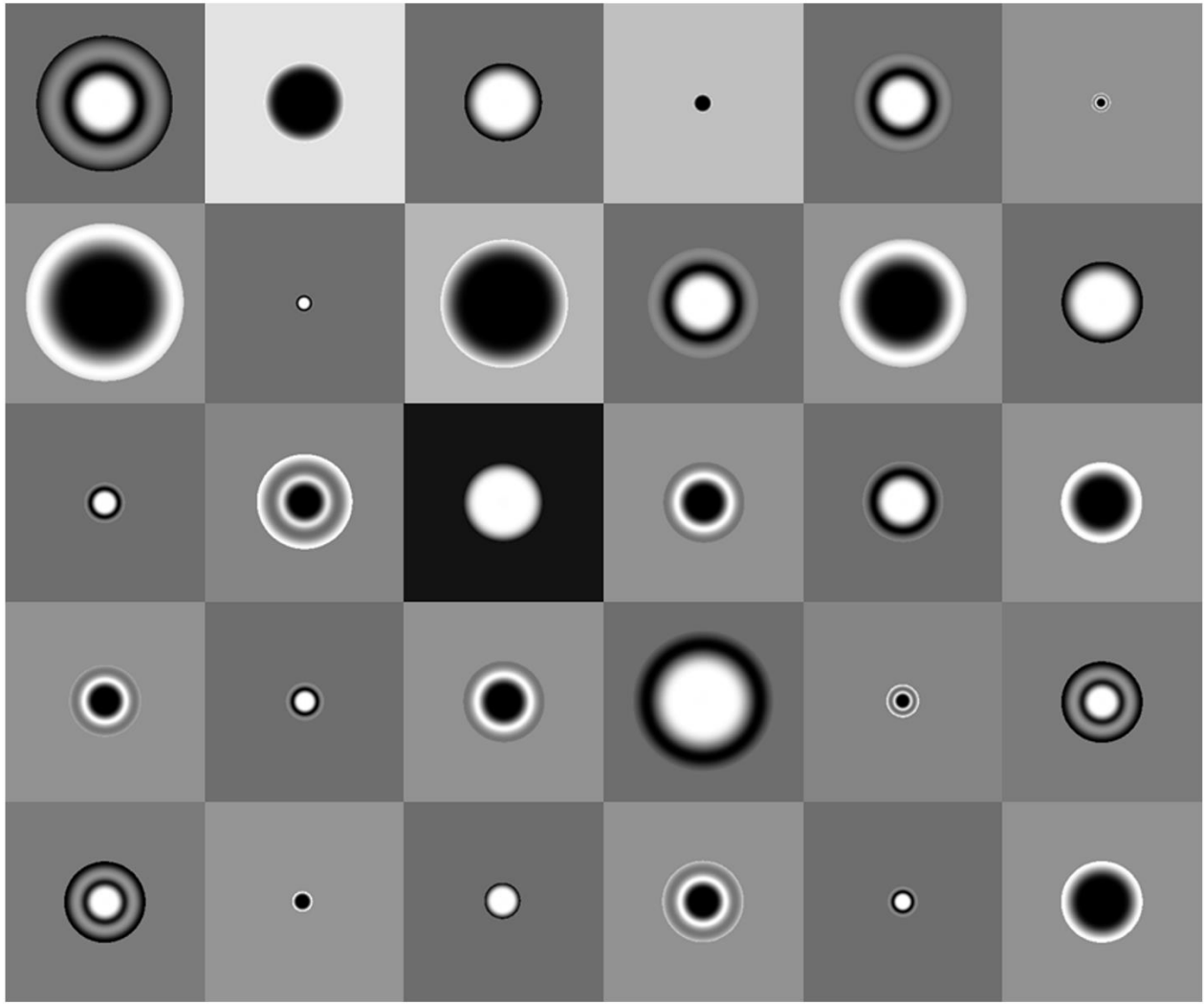

Figure S2. An example of 30 synthetic images.

## S2: Image processing techniques

Here we briefly describe the fundamentals of the algorithms compared to *C-sym*.

### 1. Center of Mass (CoM)

The center-of-mass is a simple and computational efficient algorithm (1). The center of mass position ($c_x$, $c_y$), in the X-axis and Y-axis of an image $I$, are calculated as:

$$ROI'(i,j) = \left| ROI(i,j) - I_{med} \right|, \tag{1}$$

$$c_x = \frac{\sum_{i,j} \left( i \times ROI'(i,j) \right)}{\sum_{i,j} \left( ROI'(i,j) \right)}, \tag{2}$$

$$c_y = \frac{\sum_{i,j} \left( j \times ROI'(i,j) \right)}{\sum_{i,j} \left( ROI'(i,j) \right)}, \tag{3}$$

where ($i$, $j$) are the pixel coordinates of a $n \times n$ sized ROI, centered at the initial estimation of the particle position. Since this technique is very sensitive to a non-zero background, the first step is to subtract the median $I_{med}$ of the full image from the ROI.

### 2. Circular Hough transform (CHT)

The *CHT* is a classic algorithm for finding circles in images. We implemented the *CHT* as explained in detail in (2). The algorithm works using a range of possible particle radius and the functioning of this algorithm can be summarized as follows:

**Step1:** Edge detection: A binary edge map of the image is created using a standard Sobel edge detector with a gradient threshold calculated automatically using the Otsu technique.

**Step2:** Accumulator array computation: A single 2-D accumulator array is used so pixels of high gradient are designated as being candidate pixels allowed to cast 'votes' in the accumulator array. Edge pixels cast votes for belonging to circles with a range of particle radius. Edge orientation is also used to permit voting in a limited interval along direction of the gradient.

**Step3:** The voting generates peaks in the accumulator array, and these peaks are evaluated by the algorithm as potential circles.

### 3. Cross-correlation (XCorr)

*XCorr* is a method is based on cross correlating an averaged intensity profile with its own mirror for the X and Y axes (3) (4). Based on empirical experiments, a median background subtraction process was added as a first step, similarly to the *CoM* technique. Formulation of *XCorr* can be expressed as follows:

$$ROI'(i,j) = \left| ROI(i,j) - I_{med} \right|, \tag{4}$$

$$P_X(i) = \frac{1}{0.2n} \sum_{j=0.4n}^{0.6n} ROI'(i,j), \qquad (5)$$

$$P_Y(j) = \frac{1}{0.2n} \sum_{i=0.4n}^{0.6n} ROI'(i,j), \qquad (6)$$

$$XCorr_X = \mathcal{F}^{-1}\left(\mathcal{F}\left(P_X(i)\right) \times \mathcal{F}\left(P_X^*(-i)\right)\right), \qquad (7)$$

$$XCorr_Y = \mathcal{F}^{-1}\left(\mathcal{F}\left(P_X(j)\right) \times \mathcal{F}\left(P_Y^*(-j)\right)\right), \qquad (8)$$

where $I_{med}$ is the image median subtracted to the region of interest ROI centered at the initial estimation of the particle; $P_x$ is the averaged profile of the particle, created using a band of $2n$ rows, being $n \times n$ the size of the ROI. $\mathcal{F}$ and $\mathcal{F}^{-1}$ represent the Fourier and inverse Fourier transform; and $\left(P^*(-i)\right)$ represent the conjugate transpose of profile matrix $P_x$. Here, the particle center $c_x$ is calculated finding the peak of the correlation function using a 5 point parabolic fit to obtain subpixel accuracy.

### 4. Quadrant-Interpolation (QI)

QI measurement was introduced in (4). This algorithm uses the circular geometry of the diffraction pattern to suppress bias and improve a previous measurement applied to the particle. In this work, the *XCorr* measurement is used as a first step of this technique. After this process, the functioning of the technique can be summarized as follows:

**Step1:** The position estimated by the *XCorr* technique is used to calculate a radial profile of the intensity of the particle for each quadrant on a circular grid. Here, it is assumed that the real center of the particle is within ~1 pixel of the previously estimated center. These intensity profiles are created using points spaced by δr and δq, in radial and angular dimensions δr < pixel spacing. To this end, a four neighborhood bilinear interpolation technique is used.

**Step 2:** Relative radial profiles are used to improve the estimated center. Thus, for the X-axis, an intensity profile $P_x$ is created concatenating the sum of top right and bottom right quadrant profiles with the sum of top left and bottom left ones.

$$q_R(r) = q_{TR}(r) + q_{BR}(r), \qquad (9)$$

$$q_L(r) = q_{TL}(r) + q_{BL}(r), \qquad (10)$$

$$P_x(r) = q_L(-r) \cap q_R(r), \qquad (11)$$

$$QI_x = \mathcal{F}^{-1}\left(\mathcal{F}\left(P_x(r)\right) \times \mathcal{F}\left(P_x^*(-r)\right)\right), \qquad (12)$$

where, similarly to *XCorr*, here $q_{TR}(r)$, $q_{BR}(r)$, $q_{TL}(r)$, $q_{BL}(r)$ are top right, bottom right, top left and bottom left quadrant profiles of the image; $\cap$ represents concatenation; $\mathcal{F}$ and $\mathcal{F}^{-1}$ represent the Fourier and inverse Fourier transform; and $\mathcal{F}\left(P_x^*(-i)\right)$ represent the Fourier transform of the complex conjugate. The resulting particle center $c_x$ is calculated finding the peak of the *QI* function using a five-point parabolic fit to obtain subpixel accuracy.

## S3: Central-Symmetry algorithm, median filtering and Hermite Interpolation

We evaluated the response of *C-Sym* in further detail analyzing how a median filtering and the Hermite Interpolation step affect the result of the algorithm. Median Filtering (MF) is a procedure based on replacing each pixel value of an image by its local median. MF has been widely applied to reduce the effects of noise in digital images (5). To evaluate the noise response, we compared the default version of *C-Sym* with a median filtered version. The piecewise Hermite interpolation algorithm, see Section 3, was included in the algorithm to increase the performance of *C-Sym* with moderate noise values. However, this step can be omitted to decrease computation time. To see how the Hermite interpolation affect the obtained results, we also evaluated *C-Sym* by activating and deactivating this step.

The experiment used 55 000 simulated particle images and was designed as described in section 2.1 by generating 500 simulations for each particle size and noise level. Four versions of the algorithm were therefore evaluated according to its accuracy and precision using the average and standard deviation of the Euclidean distance between the estimated position of the particle and the ground truth. The obtained results are shown in Figure S3 and Figure S4.

The results show that the Hermite interpolation step generally improves the accuracy of *C-Sym*, though this improvement is only significant in small particles. With SNR < 1, a slightly loss of accuracy may be observed and its use is thus not recommended. The use of median filter improves significantly the results with SNR < 2 with no significant change for higher SNR.

## Mean Error (pixel)

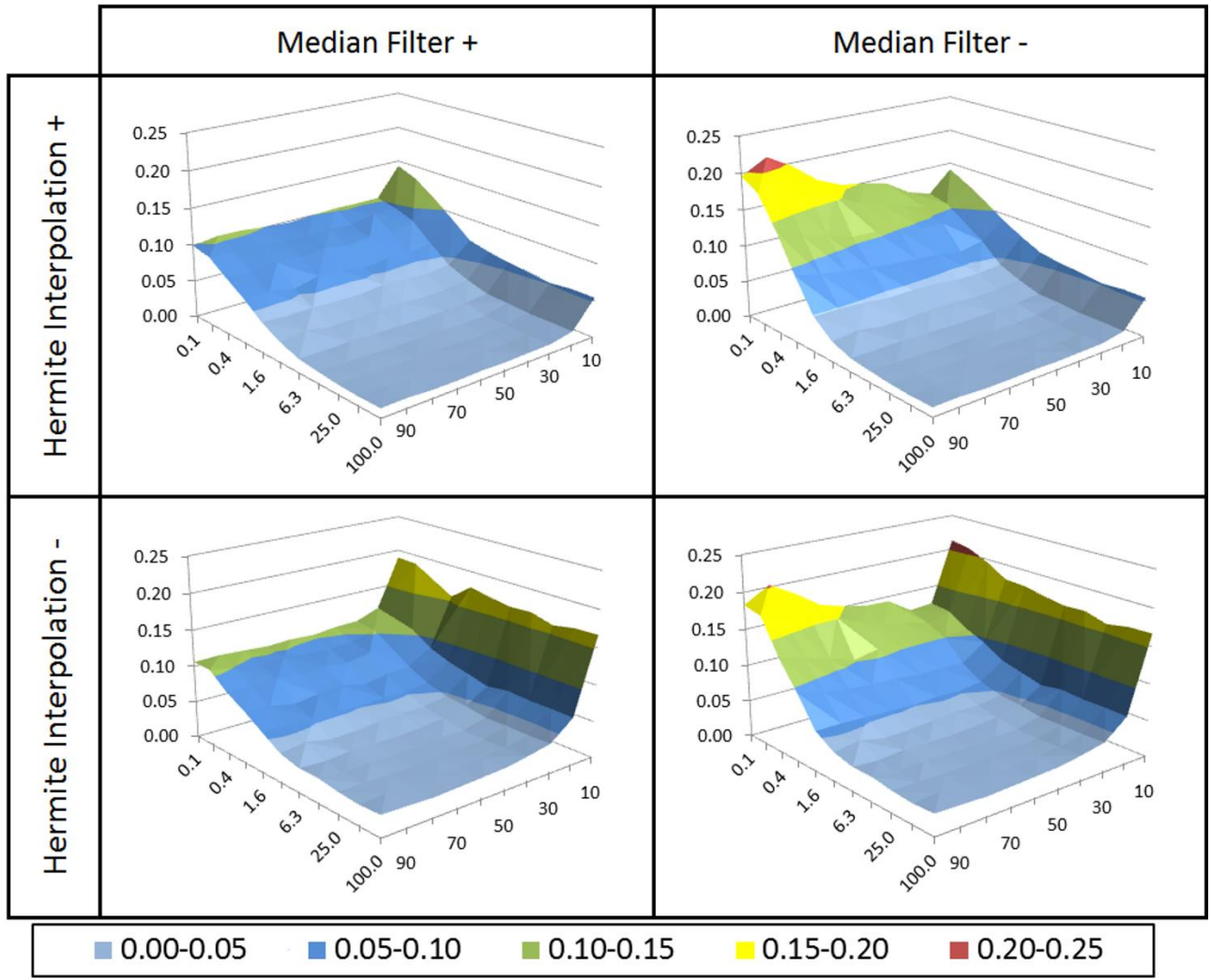

Figure S3: Mean error in the center position of the particles measured using *C-Sym*: Four different versions of the algorithm were used to study the influence in the results of a median filtering and of the Hermite Interpolation step.

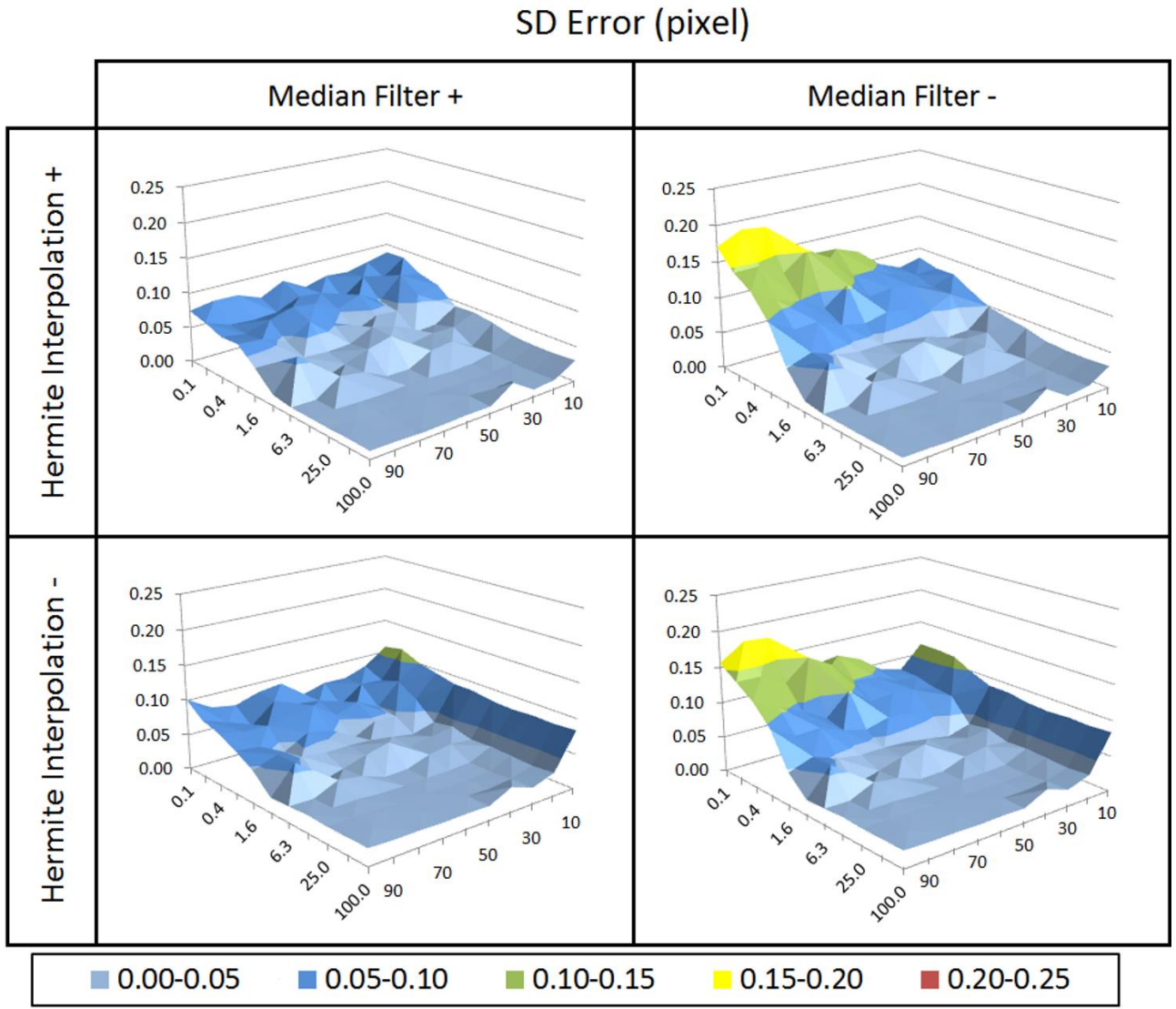

Figure S4: Standard Deviation of the error in the position of the center of the particle measured with *C-Sym*: Four versions of the algorithm were used to analyze how a median filtering and the Hermite Interpolation step influence the results.

# Supporting References